\documentclass[journal]{IEEEtran}
\usepackage{graphicx}
\usepackage[ruled,vlined]{algorithm2e}
\usepackage{multirow}
\usepackage{makecell}
\usepackage{comment}
\usepackage{amsmath,amssymb} 
\usepackage{color,colortbl}
\definecolor{LightCyan}{rgb}{0.88,1,1}

\newcommand{\eat}[1]{}

\usepackage{threeparttable}
\usepackage{enumitem}
\usepackage{multirow}
\usepackage{pifont}
\newcommand{\cmark}{\ding{51}}%
\newcommand{\xmark}{\ding{55}}%
\renewcommand\vec[1]{\overrightarrow{#1}}
\newcommand{\norm}[1]{\left\lVert#1\right\rVert}
\usepackage[colorlinks,linkcolor=blue]{hyperref}

\ifCLASSINFOpdf
\else
\fi
\begin{document}
%
\title{Anatomy-aware 3D Human Pose Estimation with Bone-based Pose Decomposition}
%
%
%
\author{Tianlang~Chen, 
        Chen Fang, 
        Xiaohui Shen, 
        Yiheng Zhu, 
        Zhili Chen, 
        and~Jiebo~Luo,~\IEEEmembership{Fellow,~IEEE}
\thanks{T. Chen, and J. Luo are with the Department of Computer Science, University of Rochester, Rochester, NY, 14627 USA (e-mail: \{zyang39, tusharku, tchen45, jluo\}@cs.rochester.edu).}
\thanks{C. Fang, X. Shen, Y. Zhu and Z. Chen are with Bytedance AI Lab, Mountain View, CA, USA (e-mail: \{fangchen, shenxiaohui, yiheng.zhu, zhili.chen\}@bytedance.com).}
\thanks{Corresponding author: Jiebo Luo.}
\thanks{Copyright\textcopyright 2021 IEEE. Personal use of this material is permitted. However, permission to use this material for any other purposes must be obtained from the IEEE by sending an email to pubs-permissions@ieee.org.}
}

\maketitle

\begin{abstract}
In this work, we propose a new solution to 3D human pose estimation in videos. Instead of directly regressing the 3D joint locations, we draw inspiration from the human skeleton anatomy and decompose the task into bone direction prediction and bone length prediction, from which the 3D joint locations can be completely derived. Our motivation is the fact that the bone lengths of a human skeleton remain consistent across time. This 
promotes us to develop effective techniques to utilize global information across {\it all} the frames in a video for high-accuracy bone length prediction. Moreover, for the bone direction prediction network, we propose a fully-convolutional propagating architecture with long skip connections. Essentially, it predicts the directions of different bones hierarchically without using any time-consuming memory units (\emph{e.g.} LSTM). A novel joint shift loss is further introduced to bridge the training of the bone length and bone direction prediction networks. Finally, we employ an implicit attention mechanism to feed the 2D keypoint visibility scores into the model as extra guidance, which significantly mitigates the depth ambiguity in many challenging poses. Our full model outperforms the previous best results on Human3.6M and MPI-INF-3DHP datasets, where comprehensive evaluation validates the effectiveness of our model.
\end{abstract}


%
\IEEEpeerreviewmaketitle

\section{Introduction}
\IEEEPARstart{3}{D} human pose estimation in videos has been widely studied in recent years. It has extensive applications in action recognition, sports analysis and human-computer interaction. Current state-of-the-art approaches \cite{pavllo20193d,lin2019trajectory,chen2019weakly} typically decompose the task into 2D keypoint detection followed by 3D pose estimation. Given an input video, they first detect the 2D keypoints of each frame, and then predict the 3D joint locations of a frame based on the 2D keypoints.

When estimating the 3D joint locations from 2D keypoints, the challenge is to resolve depth ambiguity, as multiple 3D poses with different joint depths can be projected to the same 2D keypoints. Exploiting temporal information from the video has been demonstrated to be effective for reducing this ambiguity. Typically, to predict the 3D joint locations of a frame in a video, recent approaches \cite{pavllo20193d,rayat2018exploiting,lee2018propagating} utilize temporal networks that additionally feed the adjacent frames' 2D keypoints as input. These approaches consider the adjacent local frames most associated with the current frame, and extract their information as extra guidance. However, such approaches are limited to exploiting information only from the neighboring frames
. Given a 1-minute input video with a frame rate of 50, even though we choose the existing temporal network with largest temporal window size (i.e 243 frames) \cite{pavllo20193d}, it is limited to using a concentrated short segment (about one-twelfth length of the video) to predict a single frame. Such a design can easily make existing temporal networks fail when the current frame and its adjacent input frames correspond to a complex pose, because none of the input frames provide reliable and high-confidence information to the networks. 

\vspace{-1mm}


Considering this, our first contribution is proposing a novel approach that can effectively capture the knowledge from both local and distant frames to estimate the 3D joint locations of the current frame, by cleverly exploiting the anatomic properties of the human skeleton. We refer to it as \emph{anatomy awareness}.  Specifically, based on the anatomy of the human skeleton, we decompose the task of 3D joint location prediction into two sub-tasks -- bone direction prediction and bone length prediction. We demonstrate that the combination of the two new tasks are essentially equivalent to the original task. The motivation is based on the fact that the bone lengths of a person remain consistent in a video over time (This can be verified by 3D human pose datasets such as Human3.6M and MPI-INF-3DHP). Hence, when we predict the bone lengths of a particular frame, we can leverage the frames distributed over the duration of the entire video for more accurate and smooth prediction. Note that although Sun et al. \cite{sun2017compositional} transform the task into a generic bone-based representation, such a generic representation does not allow them to utilize that critical bone length consistency.
In contrast, we decompose the task {\it explicitly into bone direction and bone length prediction}. We demonstrate that this explicit design leads to significant advantages over either the generic representation design in \cite{sun2017compositional} or imposing a bone length consistency loss across frames. 

However, it is nontrivial to implement this explicit design. One problem for training the proposed bone length prediction network is that the training dataset typically contains only a few skeletons. For example, the training set of Human3.6M contains 5 actors corresponding to 5 bone length settings. Directly training the network on the data from the 5 actors leads to serious overfitting. Therefore, we adopt the fully-connected residual network for bone length prediction and propose two effective mechanisms to prevent overfitting via a network design and data augmentation. 

As for the bone directions, we adopt the temporal convolutional network in \cite{pavllo20193d} 
to predict the direction of each bone in the 3D space for each frame. Motivated by \cite{lee2018propagating}, we believe it is beneficial to predict the directions of different bones hierarchically, instead of all at once as in \cite{pavllo20193d}. Following the human skeleton anatomy, the directions of simple torso bones (\emph{e.g.} lumbar vertebra) with less motion variation should be predicted first, and then guide the prediction of challenging limb bones (\emph{e.g.} arms and legs). This strategy is applied straightforwardly by a recurrent neural network (RNN) with different joints predicted step by step in \cite{lee2018propagating} for a single frame. However, the high computation complexity of RNN precludes the network from holding a large temporal window which has been shown to improve performance. To solve this issue, based on \cite{pavllo20193d}, we further propose a high-performance fully-convolutional propagating architecture, which contains multiple sub-networks with each predicting the directions of all the bones. The hierarchical prediction is implicitly performed via long skip connections between adjacent sub-networks. 

Additionally, motivated by \cite{sun2017compositional}, we create an effective joint shift loss for the two sub-tasks (\emph{i.e.}, bone direction prediction and bone length prediction) to learn jointly. The joint shift loss penalizes the relative joint shift between all long-range joint pairs, for example the left hand and right foot. Thus, it provides an extra strong supervision for the two networks to be trained to coordinate with each other and produce robust predictions. 


Last but not least, we propose a simple yet effective approach to further reduce the depth ambiguity. Specifically, we incorporate 2D keypoint visibility scores into the model as a new feature, which indicates the probability of each 2D keypoint being visible in a frame and provides extra knowledge of the depth relation between specific joints. We argue that the scores are useful to those poses with body parts occluded or when the relative depth matters. For example, if a person keeps her/his hands in front of the chest in a frontal view, our model will be confused on whether the hands are in front of the chest (visible) or behind the back (occluded), since  the occluded 2D keypoints can still be predicted sometimes. Furthermore, We adopt an implicit attention mechanism to dynamically adjust the importance of the visibility scores for better performance.


Our contributions are summarized as follows: 
\begin{itemize}
\vspace{-2mm}
    \item We explicitly decompose the task of 3D joint estimation into bone direction prediction and bone length prediction. As such, the bone length prediction branch can fully utilize frames across the entire video. 
    \item We propose a new fully-convolutional architecture for hierarchical bone direction prediction.
    \item We propose a high-performance bone length prediction network, two mechanisms are created to effectively prevent overfitting.
    \item We feed the visibility scores of 2D keypoint detection into the model to better resolve the depth ambiguity. 
    \item Our model is inspired by the human skeleton anatomy and achieves the state-of-the-art performance on Human3.6M and MPI-INF-3DHP datasets.
\end{itemize}

\section{Related Work}
3D human pose estimation has received much attention in recent years. To predict the 3D joint location from 2D image input, previous works of 3D pose estimation typically fall into two categories based on the training pipeline. For the approaches of the first category, they created an end-to-end convolutional neural network (CNN) model to directly predict the 3D joint location from the original input images. To establish a strong baseline, Pavlakos et al. \cite{pavlakos2017coarse} integrated the volumetric representation with a coarse-to-fine supervision scheme to figure out the 3D joint locations by the predicted 3D volumetric heat maps. Based on the ConvNet pose estimator and the volumetric heap map representation proposed by \cite{pavlakos2017coarse}, recent approaches mainly made progress from two aspects. On the one hand, human-structure constraints such as the human shape constraints \cite{zhou2017towards}, body articulation constraints \cite{yang20183d} and the joint angle constraints \cite{dabral2018learning} were employed to prevent invalid pose prediction. On the other hand, effective training approaches were proposed, making the estimation process differentiable \cite{sun2018integral} and enabling the model to learn from weakly labeled data \cite{pavlakos2018ordinal,kocabas2019self}. To further enable the pose estimator to predict full 3D human mesh instead of the joint locations, Kanazawa et al. \cite{kanazawa2018end,kanazawa2019learning} proposed end-to-end CNN frameworks for reconstructing the full 3D mesh of a human body from an image or a video. These approaches based on image-level input can directly capture rich knowledge contained in images. However, without intermediate feature and supervision, the model's performance will also be affected by the image's background, lighting and person's clothing. More importantly, the large dimension of image-level input disables the 3D model from receiving a large number of images as input, bottlenecking the performance of 3D pose estimation in video.     

For the approaches of the second category, they built a 3D joint estimation model on top of a high-performance 2D keypoint detector. Given an input image, these approaches first utilized the 2D keypoint detector to predict the image' 2D keypoints. The predicted 2D keypoints were then lifted as the 3D joint estimation model's input to predict the final 3D joint locations. As an earlier work, Chen et al. \cite{chen20173d} regarded the 3D pose estimation as a matching problem. They found the best matching 3D pose of the 2D keypoint input from the 3D pose pool by a nearest-neighbor (NN) model. Considering that the ground-truth 3D pose of the input may be non-corresponding to all the 3D poses in the pool, Martinez et al. \cite{martinez2017simple} proposed an effective fully-connected residual network to regress the 3D joint locations from 2D keypoint input. In addition to utilizing effective human-structure information as the approaches of the first category, based on \cite{martinez2017simple}, recent approaches of this category further improved the pose estimation performance by hierarchical joint prediction \cite{lee2018propagating}, 2D keypoint refinement \cite{xu2020deep} and view-invariant constraint \cite{chen2019weakly,8763991}. Overall, the approaches in such a ``image-2D-3D'' pipeline outperform the end-to-end counterparts. One important reason is that the 2D detector can be trained by large-scale indoor/outdoor images. It provides the 3D model a strong intermediate feature to build upon.

When estimating the 3D poses in a video, recent approaches exploited temporal information into the model to alleviate incoherent predictions. As an earlier work, Mehta et al. \cite{mehta2017vnect} applied simple temporal filtering across 2D and 3D poses from previous frames to predict a temporally consistent 3D pose. As Long Short Term Memory networks (LSTM) were created to adaptively capture information from temporal input by the well-designed input gate, output gate and forget gate, Lin et al. \cite{lin2017recurrent} presented the LSTM-based Recurrent 3D Pose Sequence Machine. It automatically learns the image-dependent structural constraint and sequence-dependent temporal context by a multi-stage sequential refinement. Similar to \cite{lin2017recurrent}, Rayat et al. \cite{rayat2018exploiting} predicted temporally consistent 3D poses by learning the temporal context of a sequence using sequence-to-sequence LSTM-based network. Considering the high computational complexity of LSTM,  Pavllo et al. \cite{pavllo20193d} further introduced a temporal fully-convolutional model which enables parallel processing of multiple frames and supports very long 2D keypoint sequence as input. All these approaches essentially leverage the adjacent frames to benefit the current frame's prediction. Compared with them, we are the first to make all the frames in a video contribute to the 3D prediction. Motivated by \cite{yu2017multitask,zhou2016deep,kanazawa2018end} that created effective sub-tasks for human pose estimation and mesh recovery, we propose a novel solution to decompose the 3D pose estimation task into two bone-based sub-tasks. It should be noticed that Sun et al. \cite{sun2017compositional} also transformed the 3D joint into a bone-based representation. They trained the model to regress short and long range relative shifts between different joints. We demonstrate that completely decomposing the task into the bone length and bone direction prediction achieves the best performance and makes better use of the relative joint shift supervision.  
\section{Our Model} \label{sec:model}
\vspace{-1mm}
In this section, we formally present our 3D pose estimation model. In section~\ref{sec:lengthdirection}, we first describe the overall anatomy-aware framework that decomposes the 3D joint location prediction task into bone length and direction prediction. In section~\ref{sec:longpass}, we present the fully-convolutional propagating network for hierarchical bone direction prediction. In Section~\ref{sec:length}, the architecture and training details of bone length prediction network are presented. In Section~\ref{sec:training}, we describe the framework's overall training strategy. In section~\ref{sec:att}, an implicit attention mechanism is introduced to feed the keypoint visibility scores into the model as extra guidance. Our framework's overall architecture is shown as Fig.~\ref{fig:ldframe}.
\subsection{Anatomy-aware Framework} \label{sec:lengthdirection}

\begin{figure}[!t]
\vspace{-1mm}
\centering
\includegraphics[width=3.4in]{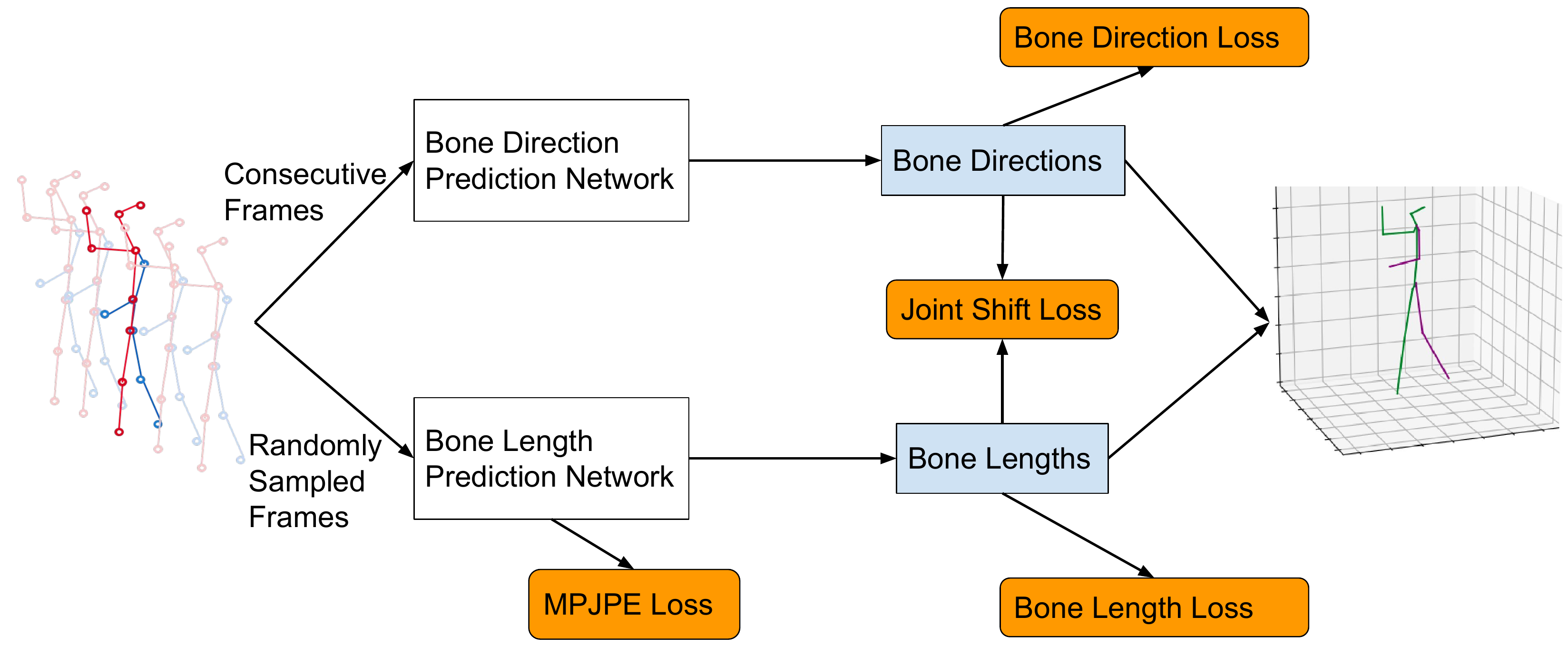}
\caption{The overview of the proposed anatomy-aware framework. It predicts the bone directions and bone lengths of the current frame using consecutive local frames and randomly sampled frames across the entire video, respectively.}
\vspace{-1mm}
\label{fig:ldframe}
\end{figure}

\begin{figure}[!t]
\vspace{-1mm}
\centering
\includegraphics[width=2.4in]{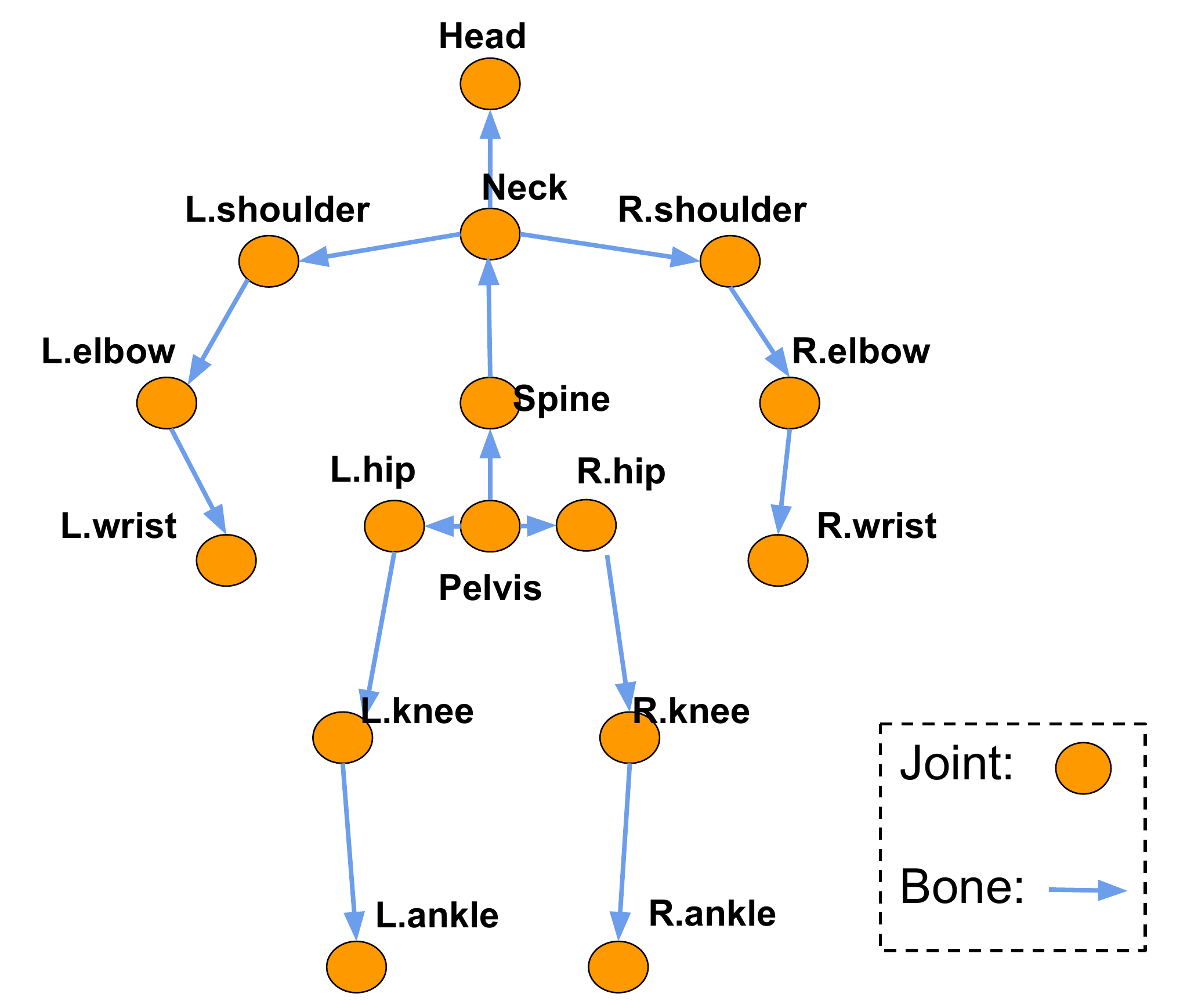}
\caption{The joint and bone representation of a human pose.}
\vspace{-1mm}
\label{fig:overview}
\end{figure}

As in \cite{pavllo20193d,rayat2018exploiting,chen2019weakly}, given the predicted 2D keypoints of each frame in a video, we aim at predicting the normalized 3D locations of $j$ pre-defined joints for each frame. The 3D location of joint ``Pelvis'' is commonly defined as the origin of the 3D coordinates. Given a human joint set that contains $j$ joints as in Fig.~\ref{fig:overview}, they correspond to $(j-1)$ directed bones with each joint being the vertex of at least one bone. This enables us to transform the 3D joint coordinates to the presentation of bone lengths and bone directions. 

Formally, to predict the 3D joint locations of a specific (\emph{i.e.} current) frame, we decompose the task to predict the length and direction of each bone. For the $k$-th joint, its 3D location $\vec{J_{k}}$ can be derived as:
\vspace{-1mm}
\begin{equation}
\begin{aligned}
\vec{J_{k}} = \sum_{b \in B^{k}}\vec{D_{b}} \cdot L_{b}
\end{aligned}
\vspace{-2mm}
\end{equation}
Here $\vec{D_{b}}$ and $L_{b}$ are the direction and length of bone $b$, respectively. $B^{k}$ contains all the bones in the path from ``Pelvis'' to the $k$-th joint.

We use two separate sub-networks to predict the bone lengths and directions of the current frame, respectively, as bone length prediction needs global input to ensure consistency across all the frames, whereas bone directions should be estimated within a local temporal window. Meanwhile, to ensure consistency between predicted bone lengths and directions, motivated by \cite{sun2017compositional}, we add a joint shift loss between the two predictions in addition to their own losses, as shown in Fig.~\ref{fig:ldframe}. Specifically, the joint shift loss is defined as follows:
\vspace{-1mm}
\begin{equation} \label{eq:2}
\begin{aligned}
\mathcal{L}_{JS} = \sum_{k_{1},k_{2} \in \mathcal{P}, k_{1}<k_{2}}\norm{X_{JS}^{k_{1}, k_{2}} - Y_{JS}^{k_{1}, k_{2}}}_{2}^{1}
\end{aligned}
\vspace{-2mm}
\end{equation}
Here $Y_{JS}^{k_{1}, k_{2}}$ is the $3$-dimensional ground-truth relative joint shift of the current frame from the $k_{1}$-th joint to the $k_{2}$-th joint, $X_{JS}^{k_{1}, k_{2}}$ is the corresponding predicted relative joint shift derived from the predicted bone lengths and bone directions of the current frame. $\mathcal{P}$ contains all the joint pairs that are not directly connected as a bone. With the joint shift loss, the two sub-networks are connected and enforced to learn from each other jointly. We describe the details of the two sub-networks in the following two sections.

\subsection{Bone Direction Prediction Network} \label{sec:longpass}
We adopt the temporal fully-convolutional network proposed by Pavllo et al. \cite{pavllo20193d} as the backbone architecture of our bone direction prediction network. Specifically, the 2D keypoints of $d$ consecutive frames are concatenated to form the input to the network, with the 2D keypoints of the current frame in the center. In essence, to predict the bone directions of the current frame, the temporal network captures the information of the current frame and the context from its adjacent frames as well. A bone direction loss based on mean squared error is applied to train the network:  
\vspace{-1mm}
\begin{equation}
\begin{aligned}
\mathcal{L}_{D} = \norm{X_{D} - Y_{D}}_{2}^{2}
\end{aligned}
\vspace{-1mm}
\end{equation}
Here $X_{D}$ and $Y_{D}$ represent the predicted and ground-truth $3(j-1)$-dimensional bone direction vector of the current frame, respectively.

\begin{figure}[!t]

\vspace{-2mm}
\centering
\includegraphics[width=3.4in]{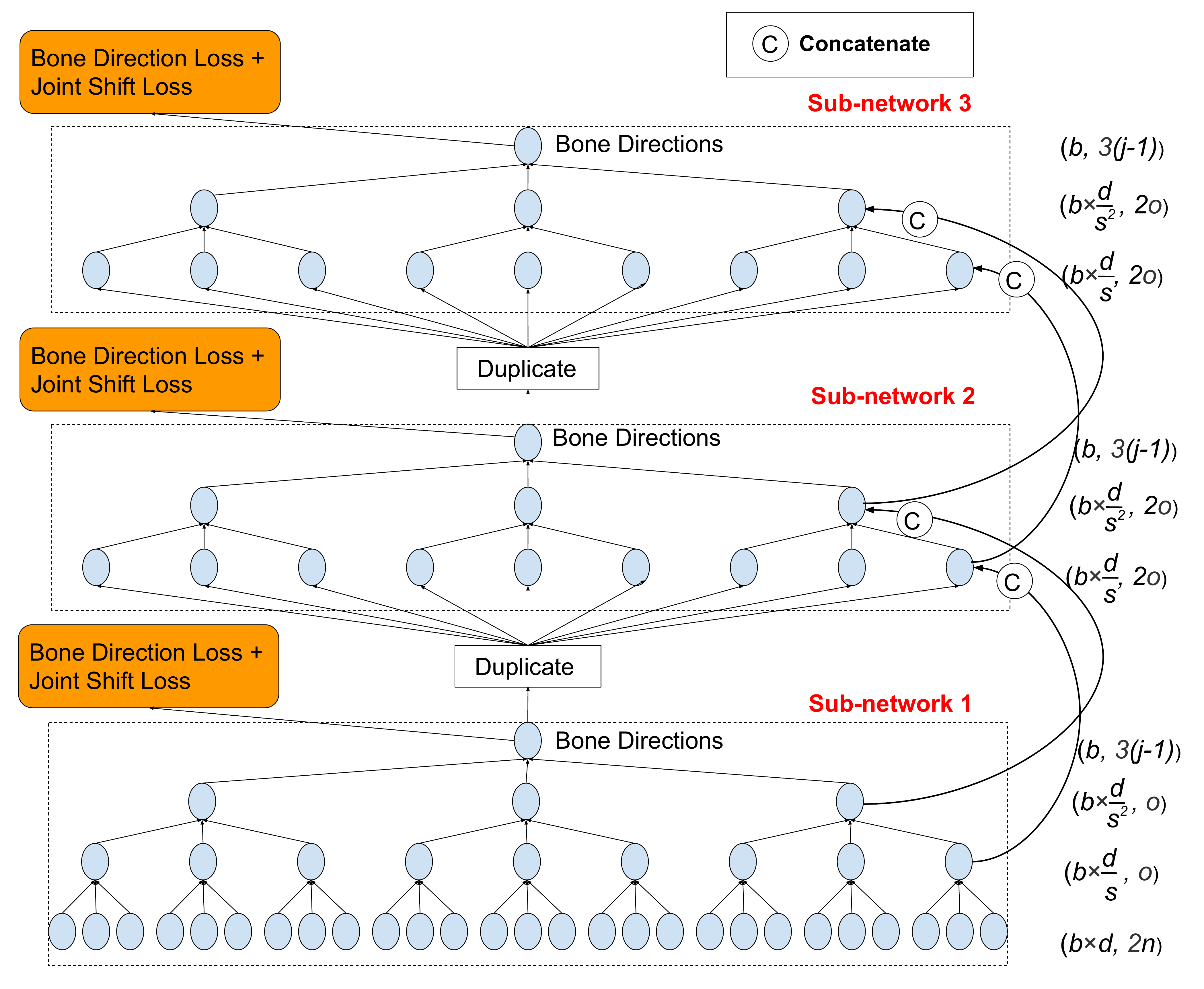}

\caption{The architecture of the bone direction prediction network. Long skip connections are added between adjacent sub-networks. We illustration the dimension of each input/output. $b$ is the batch size. $o$ is the output channel number of fully-connected layer. $s$ is the stride of 1D convolution layer in the network ($s=3$). $n$ is the size of the 2D keypoint set. $d$ is
the input frame number of the bottom sub-network.}
\vspace{-1mm}
\label{fig:longpass}
\end{figure}

It should be noted that the joint shift loss introduced in Section~\ref{sec:lengthdirection} makes the predicted directions of different bones mutually relevant. For example, if the predicted direction of the left lower arm is inaccurate, the predicted direction of the left upper arm will also be affected, since the model is encouraged to regress a long range shift from left shoulder to left wrist. Intuitively, it would benefit the overall prediction if we could first predict those easy and high-confident cases, and let them guide the subsequent prediction of other joints. As poses may vary significantly, it is difficult to pre-determine the hierarchy of the prediction. Motivated by \cite{lee2018propagating}, here we propose a fully-convolutional propagating architecture with long skip connections, and let the network itself to learn the prediction hierarchy instead, as in Fig.~\ref{fig:longpass}.

\begin{figure}[!t]

\vspace{+0mm}
\centering
\includegraphics[width=3.2in]{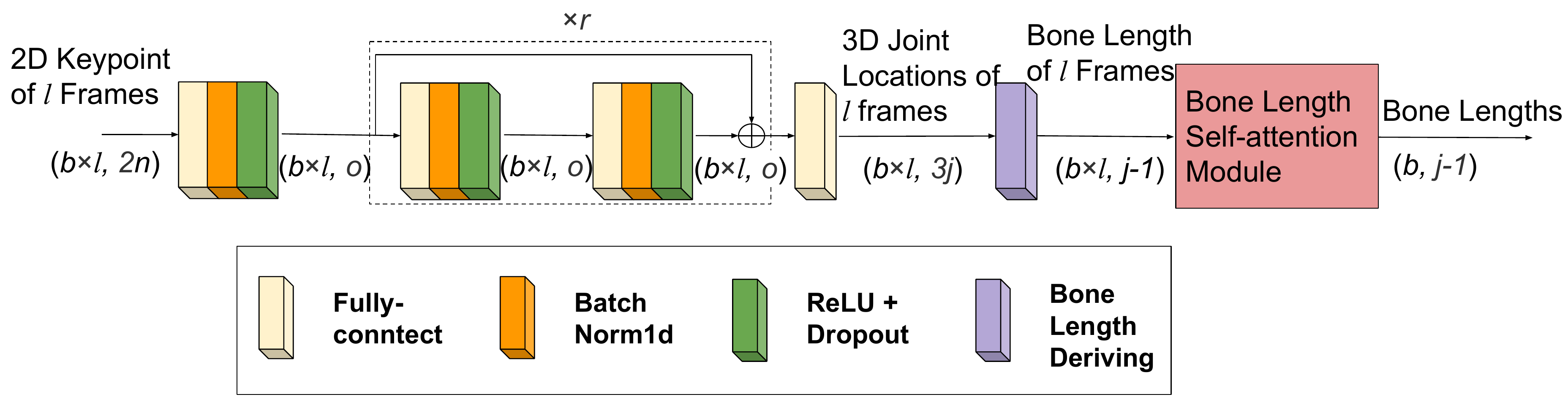}

\caption{Detailed structure of the bone length prediction network. $r$ is the number of residual block.}
\vspace{-1mm}
\label{fig:bl}
\end{figure}

Specifically, the architecture is a stack of several sub-networks, with each sub-network being a temporal fully-convolutional network with residual blocks proposed by \cite{pavllo20193d}. The output of each sub-network is the predicted bone directions of the current frame. Except the top sub-network, we temporally duplicate the output of each sub-network $\frac{d}{s}$ times as the input to the next sub-network. For each residual block of a specific sub-network, we concatenate its output with the output of the corresponding residual block in the adjacent upper sub-network on channel level. This forms the long skip connections between adjacent sub-networks. We adopt an independent training strategy for each sub-network, that is, we train each sub-network by the loss of the bone direction prediction network, the back propagation is blocked between different sub-networks. By doing that, the bottom networks would not be affected by the upper ones, and instead would propagate high-confident predictions to guide subsequent predictions. In the process, the model automatically learns the hierarchical order of the prediction. In Section~\ref{sec:exp}, we demonstrate the effectiveness of the proposed architecture.


\subsection{Bone Length Prediction Network} \label{sec:length}

As discussed in Section~\ref{sec:lengthdirection}, the prediction of bone lengths requires global inputs from the entire video. However, taking too many frames as the input would make the computation prohibitively expensive. To capture the global context efficiently, we choose to randomly sample $l$ frames across the entire video as the input to the network. The detailed structure of the network is shown as Fig.~\ref{fig:bl}.

We adopt the fully-connected residual network for bone length prediction. Specifically, it has the same structure and layer number as the bottom sub-network of the bone direction prediction network. However, since the randomly sampled frames do not have temporal connections, we replace each 1D convolution layer by the fully-connected layer in the network. This adapts the network for single-frame input instead of multi-frame consecutive inputs. The fully-connected network predicts the $(j-1)$ bone lengths of each sampled frame.

Intuitively, we can average the predicted bone lengths of each sampled frame as the 
predicted bone lengths of the current frame. In such a way, a similar bone length loss can be applied to train the fully-connected network:  
\vspace{-1mm}
\begin{equation} \label{equ:ll}
\begin{aligned}
\mathcal{L}_{L} = \norm{X_{L} - Y_{L}}_{2}^{2}
\end{aligned}
\vspace{-1mm}
\end{equation}
Here $X_{L}$ and $Y_{L}$ are the predicted and ground-truth $(j-1)$-dimensional bone length vector of the current frame.

However, since the training datasets usually only contain very limited number of actors, and the bone lengths in the videos performed by the same actor are identical. Such a training loss would lead to severe overfitting. To solve this problem, instead of directly predicting the bone lengths, the fully-connected residual network is modified to predict the 3D joint locations of each sampled frame, supervised by the mean per joint position error (MPJPE) loss as in \cite{pavllo20193d}:
\vspace{-1mm} 
\begin{equation} \label{equ:lj}
\begin{aligned}
\mathcal{L}_{J} = 
\frac{1}{j}\sum_{k=1}^j\norm{X_{J}^{k} - Y_{J}^{k}}_{2}^{1}\\
\end{aligned}
\vspace{-1mm}
\end{equation}
Here $X_{J}^{k}$ and $Y_{J}^{k}$ are the predicted and ground-truth $3$-dimensional 3D joint locations of the $k$-th joint. Since each frame would predict a set of 3D joint locations, and the number of input frames $l$ is usually large, minimizing the MPJPE for the predictions of all the frames would make the convergence speed of the bone length prediction network much faster than the bone direction prediction network. This decreases the performance of jointly training them by Equation~\ref{eq:2}.
We choose to randomly sample one frame from $l$ input frames and calculate its corresponding MPJPE loss. We find that such a training strategy works quite well as shown in the experiments. 

Since the 3D joint locations would vary in each frame, the overfitting problem would be largely avoided.  The bone lengths of each of the $l$ frame can then be derived from the 3D joint locations accordingly.

When averaging the bone length predictions from the input frames, the prediction accuracy for a specific bone depends on the poses. For example, some of the bones in a certain pose are hardly visible due to occlusion or foreshortening, making the prediction unreliable. Motivated by the self-attention mechanism that is widely used for vision and language tasks \cite{vaswani2017attention,yu2019multimodal,yu2020deep,li2019beyond,8920126}, we further incorporate a self-attention module at the top of the fully-connected network to predict the bone length vector $X_{L}$ of the current frame:
\vspace{-1mm}
\begin{equation}
\label{eq:6}
\begin{aligned}
A_{i} & = \frac{\exp(\gamma WX_{iJ})}{\sum_{i=1}^{l}\exp(\gamma WX_{iJ})}\\
X_{L}  & = \sum_{i=1}^l A_{i} \odot  X_{iL}\\
\end{aligned}
\vspace{-1mm}
\end{equation}
Here $X_{iJ}$ and $X_{iL}$ are the predicted $3j$-dimentional 3D joint location vector and the corresponding derived bone length vector of the $i$-th input frame. $W$ is a learnable matrix of the self-attention module. $\odot$ indicates element-wise multiplication. $A_{i}$ is the $(j-1)$-dimensional attention weights that indicate the bone-specific importance of the $i$-th frame's predicted bone lengths for $X_{L}$. $\gamma$ is a hyper-parameter to control the degree of attention. During training, the fully-connected residual network and the bone length self-attention module are optimized independently by $L_{J}$ (Equation~\ref{equ:lj}) and $L_{L}$ (Equation~\ref{equ:ll}). 

To further solve the overfitting problem of the self-attention module, we augment the training data by generating samples with variant bone lengths. In particular, for each training video, we randomly create a new group of $(j-1)$ bone lengths. We modify the ground-truth 3D joint locations of each frame to make them accordant with the new bone lengths. Because the camera parameter is available, we can reconstruct the 2D keypoints of each frame from its modified ground-truth 3D joint locations. For each training iteration, we additionally feed a batch of randomly sampled $l$ frames from the augmented videos and use the corresponding 2D keypoints and bone lengths to optimize the self-attention module by $\mathcal{L}_{L}$ . As the self-attention module is only used for bone length re-weighting, we consider it valid to train this module by a combination of predicted 2D keypoints and reconstructed clean 2D keypoints.  

\subsection{Overall Loss Function} \label{sec:training}
By combining the losses in each sub-network and the joint shift loss, the overall loss function for our framework is given as:
\vspace{-1mm}
\begin{equation}
\label{eq:7}
\begin{aligned}
\mathcal{L} = \lambda_{D}{\mathcal{L}_{D}} + \lambda_{L}{\mathcal{L}_{L}} + \lambda_{J}{\mathcal{L}_{J}} + \lambda_{JS}{\mathcal{L}_{JS}}
\end{aligned}
\vspace{-1mm}
\end{equation}
Here $\lambda_{D}$, $\lambda_{L}$, $\lambda_{J}$ and $\lambda_{JS}$ are hyper-parameters regulating the importance of each term.

During training, only the parameters of the bone direction prediction network are updated by the joint shift loss. Essentially, the joint shift loss supervises the model to predict robust bone directions that match with the predicted bone lengths for long range objectives. In Section~\ref{sec:exp}, we prove that the proposed anatomy-aware framework better exerts the joint shift loss's potential than \cite{sun2017compositional}.

During inference, to estimate the 3D joint locations of a specific frame, we adopt the same strategy as the training process. We still randomly sample $l$ frames of the video to predict the bone lengths of this frame. We find that taking all the frames of the video as input for bone length prediction does not lead to better performance. In Section~\ref{sec:exp}, we provided more details regarding the frame sampling strategy for bone length prediction.

\subsection{Incorporating the Visibility Scores} \label{sec:att}
\begin{figure}[!t]
\vspace{-2mm}
\centering
\includegraphics[width=3.1in]{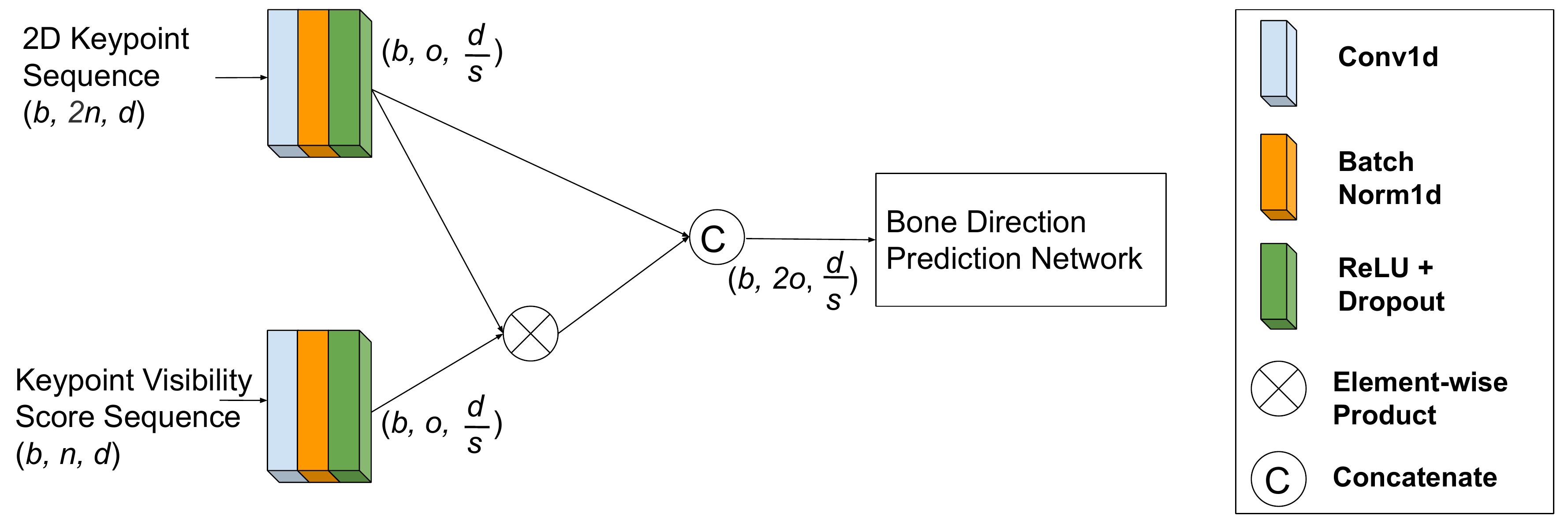}

\caption{Detailed network structure for feeding the visibility scores. $d$ is the input frame number of the bone direction prediction network, $s$ and $o$ are the stride and output channel number of 1D convolution layer, respectively.}
\vspace{-1mm}
\label{fig:vs}
\end{figure}

Our 3D joint prediction network takes the predicted 2D keypoints as the input, which sometimes have ambiguities. For example, when a person has his/her legs crossed in a front view, the corresponding 2D keypoints cannot provide information about which leg is in the front. A more common situation happens when the person put his/her hands in front of the chest or behind the back, the 3D model will be confused by the relative depth between the hands and chest. Even though temporal information is exploited, the above problems still exist. 

We provide a simple yet effective approach to solve the problem without feeding the original images into the model. Specifically, we predict the visibility score of each 2D keypoint in a frame, and incorporate it into the model for 3D joint estimation. The visibility score indicates the confidence of each keypoint being visible in the frame, which can be extracted from most 2D keypoint detectors. Compared with the position, a keypoint's visibility is easy to be predicted, this score is thus reliable.

We argue that the importance of a specific keypoint's visibility score is related to the corresponding pose. For example, the visibility scores of the hands become useless if the hands are stretched far away from the body. Therefore, we adopt an implicit attention mechanism to adaptively adjust the importance of the visibility scores.


Given the 2D keypoint sequence of $d$ consecutive frames as the input of the network in Section~\ref{sec:longpass}, we feed the keypoint visibility score sequence of the $d$ frames as in Fig.~\ref{fig:vs}. An 1D convolutional block is first applied, which maps the visibility score sequence into a temporal hidden feature that has the same dimension as the hidden feature of the 2D keypoint sequence. After that, we do element-wise multiplication for the two temporal hidden features as the weighted visibility score feature. In the end, we concatenate the weighted visibility score feature and the hidden feature of the 2D keypoint sequence on channel level and feed the concatenated feature to the next 1D convolution layer of the network. The whole process can be expressed as follows:

\begin{equation}
\label{eq:6}
\begin{aligned}
O_{I} = [C_{1D}(X_{I}) \odot C_{1D}(V_{I}); C_{1D}(X_{I})]   
\end{aligned}
\vspace{-1mm}
\end{equation}

Here $X_{I}$, $V_{I}$ and $O_{I}$ represent the input 2D keypoint sequence, the input keypoint visibility score sequence and the outputted concatenated features. $C_{1D}$ represent the operations performed by a combination of 1D convolution layer, batch normalization layer and ReLU unit.$[\cdot; \cdot]$ indicates concatenation. Similar to \cite{kim2016multimodal}, the hidden feature of the 2D keypoint sequence can be regarded as implicit attention weights to adjust the visibility score importance. 

\section{Experiments} \label{sec:exp}
\subsection{Datasets and Evaluation}
We evaluate the proposed model on two well established 3D human pose estimation datasets: Human3.6M \cite{ionescu2013human3} and MPI-INF-3DHP \cite{mehta2017monocular}.
\begin{itemize}[leftmargin=0.2in]
\setlength\itemsep{0.0em}

\item \textbf{Human3.6M} contains 3.6 million video frames with the corresponding annotated 3D and 2D human joint positions, from 11 actors. Each actor performs 15 different activities captured from 4 camera views. 
Following previous works \cite{pavllo20193d,lee2018propagating,rayat2018exploiting,sun2017compositional,martinez2017simple}, the model is trained on five subjects (S1, S5, S6, S7, S8) and evaluated
on two subjects (S9 and S11) on a 17-joint skeleton. We follow the standard protocols to evaluate the models on Human3.6M. The first one (\emph{i.e.} Protocol 1) is the mean per-joint position error (MPJPE) in millimeters that measures the mean Euclidean distance between the predicted and ground-truth joint positions without any transformation. The second one (\emph{i.e.} Protocol 2) is the normalized variant P-MPJPE after aligning the predicted 3D pose with the ground-truth using a similarity transformation. In addition, to measure the smoothness
of predictions over time, which is important for video, we also report the joint velocity errors (MPJVE) created by \cite{pavllo20193d} corresponding to the MPJPE of the first derivative of the
3D pose sequences.

\item \textbf{MPI-INF-3DHP} is a recently proposed 3D dataset consisting of both constrained indoor and complex outdoor scenes. It records 8 actors performing 8 activities from 14 camera views. Following \cite{mehta2017vnect,mehta2017monocular}, on a 14-joint skeleton, we consider all the 8 actors in the training set and select sequences from 8 camera views in total (5 chest-high cameras, 2 head-high cameras and 1 knee-high camera) for training. Evaluation is performed on the independent MPI-INF-3DHP test set that has different scenes, camera views and relatively different actions from
the training set. This design implicitly covers the cross-dataset evaluation. We report the Percentage of Correct Keypoints (PCK) within 150mm range, Area Under Curve (AUC), and MPJPE. 

\end{itemize}

\subsection{Implementation details} \label{sec:id}

For Human3.6M, we use the predicted
2D keypoints released by \cite{pavllo20193d} from the Cascaded Pyramid Network
(CPN) as the input of our 3D pose model. For MPI-INF-3DHP, the predicted 2D keypoints are acquired from the pretrained AlphaPose model \cite{fang2017rmpe}. 
In addition to the 2D keypoints, the keypoint visibility scores for both datasets are also extracted from the pretrained AlphaPose model. 

We use the Adam optimizer to train our model in an end-to-end manner. For each training iteration, the mini-batch size is set to 1024 for both original samples and augmented samples. We set $\lambda_{D}$ = 0.02, $\lambda_{L}$ = 0.05, $\lambda_{J}$ = 1 and $\lambda_{JS}$ = 0.1 for the loss terms in Equation \ref{eq:7}. For the bone length self-attention module, we set $\gamma$ = 10 in Equation \ref{eq:6}. The sampled frame number of the bone length prediction network $l$ is set to 50 for both the training and inference process. For the proposed architecture in Section~\ref{sec:longpass}, the number of sub-networks is set to 2. As in \cite{pavllo20193d}, the output channel number of each 1D convolution layer and fully-connected layer is set to 1024. For actual implementation, instead of manually deriving the 3D joint locations and relative joint shifts from the predicted bone lengths and bone directions, we regress the two objectives by feeding the concatenation of the predicted bone length vector and bone direction vector into two fully-connected layers, respectively. The fully-connected layers are trained together with the whole network. This achieves slightly better performance.  

\vspace{-1mm}
\subsection{Experiment results} \label{sec:expr}

\begin{table*}[htbp]
  \vspace{-2mm}
  \small\centering
  \caption{\label{tab:result1} Quantitative comparisons between the estimated pose and the ground-truth on Human3.6M under Protocols 1,2 and MPJVE. (*) We report the result without data augmentation using virtual cameras.}
  \begin{threeparttable}
  \scalebox{0.62}{
  \begin{tabular}{|c|c|c|c|c|c|c|c|c|c|c|c|c|c|c|c|c|}

    \cline{1-17}
    
    Protocol 1&Dir.&Disc.&Eat&Greet&Phone&Photo&Pose&Purch.&Sit&SitD.&Smoke&Wait&WalkD.&Walk&WalkT.&Avg \\ 
    Martinez et al. \cite{martinez2017simple} ICCV'17 &51.8 &56.2 &58.1 &59.0 &69.5 &78.4 &55.2 &58.1 &74.0 &94.6 &62.3 &59.1 &65.1 &49.5 &52.4 &62.9 \\ 
    Sun et al. \cite{sun2017compositional} ICCV’17 &52.8 &54.8 &54.2 &54.3 &61.8 &67.2 &53.1 &53.6 &71.7 &86.7 &61.5 &53.4 &61.6 &47.1 &53.4 &59.1 \\ 
    Pavlakos et al. \cite{pavlakos2018ordinal} CVPR’18 &48.5 &54.4 &54.4 &52.0 &59.4 &65.3 &49.9 &52.9 &65.8 &71.1 &56.6 &52.9 &60.9 &44.7 &47.8 &56.2 \\
    Yang et al. \cite{yang20183d} CVPR’18 &51.5 &58.9 &50.4 &57.0 &62.1 &65.4 &49.8 &52.7 &69.2 &85.2 &57.4 &58.4 &43.6 &60.1 &47.7 &58.6 \\
    Luvizon et al. \cite{luvizon20182d} CVPR’18 &49.2 &51.6 &47.6 &50.5 &51.8 &60.3 &48.5 &51.7 &61.5 &70.9 &53.7 &48.9 &57.9 &44.4 &48.9 &53.2 \\
    Hossain \& Little \cite{rayat2018exploiting} ECCV’18 &48.4 &50.7 &57.2 &55.2 &63.1 &72.6 &53.0 &51.7 &66.1 &80.9 &59.0 &57.3 &62.4 &46.6 &49.6 &58.3 \\
    Lee et al. \cite{lee2018propagating} ECCV’18 &\textbf{40.2} &49.2 &47.8 &52.6 &50.1 &75.0 &50.2 &43.0 &55.8 &73.9 &54.1 &55.6 &58.2 &43.3 &43.3 &52.8 \\
    Chen et al. \cite{chen2019weakly} CVPR'19 &41.1 &44.2 &44.9 &45.9 &46.5 &\textbf{39.3} &\textbf{41.6} &54.8 &73.2 &\textbf{46.2} &48.7 &42.1 &\textbf{35.8} &46.6 &38.5 &46.3 \\
    Pavllo et al. \cite{pavllo20193d} (243 frames, Causal) CVPR'19 &45.9&48.5&44.3&47.8&51.9&57.8&46.2&45.6&59.9&68.5&50.6&46.4&51.0&34.5&35.4&49.0\\
    Pavllo et al. \cite{pavllo20193d} (243 frames) CVPR'19 &45.2 &46.7 &43.3 &45.6 &48.1 &55.1 &44.6 &44.3 &57.3 &65.8 &47.1 &44.0 &49.0 &32.8 &33.9 &46.8\\
    Lin et al. \cite{lin2019trajectory} BMVC'19 &42.5&44.8&42.6&44.2&48.5&57.1&42.6&\textbf{41.4}&56.5&64.5&47.4&43.0&48.1&33.0&35.1&46.6\\
    
    Cai et al. \cite{cai2019exploiting} ICCV'19 &44.6 &47.4 &45.6 &48.8 &50.8 &59.0 &47.2 &43.9 &57.9 &61.9 &49.7 &46.6 &51.3 &37.1 &39.4 &48.8\\
    Cheng et al. \cite{cheng2019occlusion} ICCV'19 (*) &- &- &- &- &- &- &- &- &- &- &- &- &- &- &- &44.8\\
    Yeh et al. \cite{yeh2019chirality} NIPS'19 &44.8 &46.1 &43.3 &46.4 &49.0 &55.2 &44.6 &44.0 &58.3 &62.7 &47.1 &43.9 &48.6 &32.7 &33.3 &46.7\\
    Xu et al. \cite{xu2020deep} CVPR'20 &37.4 &43.5 &42.7 &42.7 &46.6 &59.7 &41.3 &45.1 &52.7 &60.2 &45.8 &43.1 &47.7 &33.7 &37.1 &45.6 \\
    \cline{1-17}
    Ours (243 frames, Causal) &42.5&45.4&42.3&45.2&49.1&56.1&43.8&44.9&56.3&64.3&47.9&43.6&48.1&34.3&35.2&46.6\\
    Ours (81 frames) &42.1 &43.8 &41.0 &43.8 &\textbf{46.1} &53.5 &42.4 &43.1 &\textbf{53.9} &60.5 &45.7 &42.1 &46.2 &32.2 &33.8 &44.6 \\
    Ours (243 frames) &41.4 &\textbf{43.5} &\textbf{40.1} &\textbf{42.9} &46.6 &51.9 &41.7 &42.3 &\textbf{53.9} &60.2 &\textbf{45.4} &\textbf{41.7} &46.0 &\textbf{31.5} &\textbf{32.7} &\textbf{44.1} \\ \cline{1-17}

    Protocol 2&Dir.&Disc.&Eat&Greet&Phone&Photo&Pose&Purch.&Sit&SitD.&Smoke&Wait&WalkD.&Walk&WalkT.&Avg \\
    Martinez et al. \cite{martinez2017simple} ICCV'17 &39.5 &43.2 &46.4 &47.0 &51.0 &56.0 &41.4 &40.6 &56.5 &69.4 &49.2 &45.0 &49.5 &38.0 &43.1 &47.7 \\
    Sun et al. \cite{sun2017compositional} ICCV'17 &42.1 &44.3 &45.0 &45.4 &51.5 &53.0 &43.2 &41.3 &59.3 &73.3 &51.0 &44.0 &48.0 &38.3 &44.8 &48.3 \\
    Pavlakos et al. \cite{pavlakos2018ordinal} CVPR'18 &34.7 &39.8 &41.8 &38.6 &42.5 &47.5 &38.0 &36.6 &50.7 &56.8 &42.6 &39.6 &43.9 &32.1 &36.5 &41.8 \\
    Yang et al. \cite{yang20183d} CVPR’18 &\textbf{26.9} &\textbf{30.9} &36.3 &39.9 &43.9 &47.4 &\textbf{28.8} &\textbf{29.4} &\textbf{36.9} &58.4 &41.5 &\textbf{30.5} &\textbf{29.5} &42.5 &32.2 &37.7 \\
    Hossain \& Little \cite{rayat2018exploiting} ECCV'18 &35.7 &39.3 &44.6 &43.0 &47.2 &54.0 &38.3 &37.5 &51.6 &61.3 &46.5 &41.4 &47.3 &34.2 &39.4 &44.1 \\
    Chen et al. \cite{chen2019weakly} CVPR'19 &36.9 &39.3 &40.5 &41.2 &42.0 &\textbf{34.9} &38.0 &51.2 &67.5 &\textbf{42.1} &42.5 &37.5 &30.6 &40.2 &34.2 &41.6 \\
    Pavllo et al. \cite{pavllo20193d} (243 frames, Causal) CVPR'19 &35.1&37.7&36.1&38.8&38.5&44.7&35.4&34.7&46.7&53.9&39.6&35.4&39.4&27.3&28.6&38.1\\
    Pavllo et al. \cite{pavllo20193d} (243 frames) CVPR'19 &34.1 &36.1 &34.4 &37.2 &36.4 &42.2 &34.4 &33.6 &45.0 &52.5 &37.4 &33.8 &37.8 &25.6 &27.3 &36.5\\ 
    Lin et al. \cite{lin2019trajectory} BMVC'19 &32.5&35.3&34.3&36.2&37.8&43.0&33.0&32.2&45.7&51.8&38.4&32.8&37.5&25.8&28.9&36.8\\
    Cai et al. \cite{cai2019exploiting} ICCV'19 &35.7 &37.8 &36.9 &40.7 &39.6 &45.2 &37.4 &34.5 &46.9 &50.1 &40.5 &36.1 &41.0 &29.6 &33.2 &39.0\\
    Cheng et al. \cite{cheng2019occlusion} ICCV'19 (*) &- &- &- &- &- &- &- &- &- &- &- &- &- &- &- &\textbf{34.1}\\
    Xu et al. \cite{xu2020deep} CVPR'20 &31.0 &34.8 &34.7 &34.4 &36.2 &43.9 &31.6 &33.5 &42.3 &49.0 &37.1 &33.0 &39.1 &26.9 &31.9 &36.2 \\
    \cline{1-17}
    Ours (243 frames, Causal) &33.6&36.0&34.4&36.6&37.5&42.6&33.5&33.8&44.4&51.0&38.3&33.6&37.7&26.7&28.2&36.5\\
    Ours (81 frames) &33.1 &35.3 &33.4 &35.9 &\textbf{36.1} &41.7 &32.8 &33.3 &42.6 &49.4 &37.0 &32.7 &36.5 &25.5 &27.9 &35.6 \\
    Ours (243 frames) &32.6 &35.1 &\textbf{32.8} &\textbf{35.4} &36.3 &40.4 &32.4 &32.3 &42.7 &49.0 &\textbf{36.8} &32.4 &36.0 &\textbf{24.9} &\textbf{26.5} &35.0 \\ \cline{1-17}
    
    MPJVE&Dir.&Disc.&Eat&Greet&Phone&Photo&Pose&Purch.&Sit&SitD.&Smoke&Wait&WalkD.&Walk&WalkT.&Avg \\

    Pavllo et al. \cite{pavllo20193d} (243 frames) CVPR'19 &3.0 &3.1 &2.2 &3.4 &2.3 &2.7 &2.7 &3.1 &2.1 &2.9 &2.3 &2.4 &3.7 &3.1 &2.8 &2.8\\
    Ours (243 frames) &\textbf{2.7}&\textbf{2.8}&\textbf{2.0}&\textbf{3.1}&\textbf{2.0}&\textbf{2.4}&\textbf{2.4}&\textbf{2.8}&\textbf{1.8}&\textbf{2.4}&\textbf{2.0}&\textbf{2.1}&\textbf{3.4}&\textbf{2.7}&\textbf{2.4}&\textbf{2.5}\\ \cline{1-17}

    \end{tabular}}%
  \end{threeparttable}
  
  \vspace{-1mm}
\end{table*}%
\vspace{-0mm}

\begin{figure*}[!t]
\vspace{-1mm}
\centering
\includegraphics[width=6in]{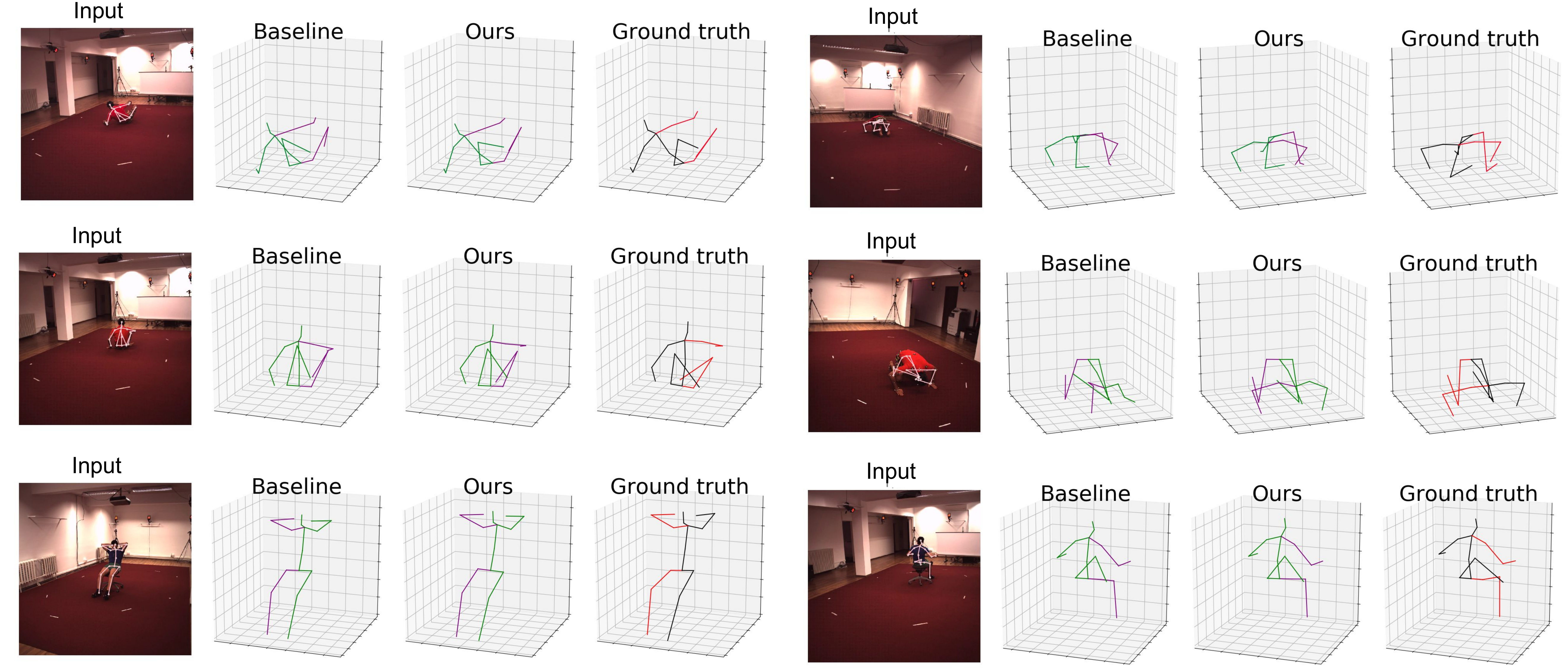}
\caption{Qualitative comparison between the proposed 243-frame model and the baseline 243-frame model \cite{pavllo20193d} on typical poses.}
\label{fig:compare}
\vspace{-1mm}
\end{figure*}

Table~\ref{tab:result1} shows the quantitative results of our proposed full model and other baselines on Human3.6M. Following  \cite{pavllo20193d}, we present the performance of our 81-frame and 243-frame models which receive 81 and 243 consecutive frames, respectively, as the input of the bone direction prediction network. We also experiment with a causal version of our model to enable real-time prediction. During the training/inference process, the causal model only receives $d$ consecutive and $l$ randomly sampled frames from the past/current frames for the current frame's estimation. Overall, our model has low average error on both Protocol 1, Protocol 2 and MPJVE. On a great number of actions, we achieve the best performance. Compared with the baseline model  \cite{pavllo20193d} that shares the same 2D keypoint detector, our model achieves more smooth prediction with lower MPJVE and achieves significantly better performance on complex activities such as ``Sitting'' (-3.4mm in Protocol 1) and ``Sitting down'' (-5.6mm in Protocol 1). We attribute it to the accurate prediction of the bone lengths for these activities. Even though the person bends his/her body, based on the predicted bone lengths, the joint shift loss can effectively guide the model to predict high-quality bone directions. Fig.~\ref{fig:compare} shows the visualized qualitative results from the baseline and our full model on ``Sitting'' and ``Sitting down'' poses.

\begin{table}[htbp]

  \small\centering
  \caption{\label{tab:result2} Quantitative comparisons of models trained/evaluated on Human3.6M using the ground-truth 2D input.}
  \begin{threeparttable}
  \scalebox{0.8}{
  \begin{tabular}{|c|c|c|}
  \cline{1-3}
  & Protocol 1 & Protocol 2\\\cline{1-3}
  Martinez et al. \cite{martinez2017simple} &45.5 & 37.1 \\
  Hossain \& Little \cite{rayat2018exploiting} &41.6 &31.7  \\
  Lee et al. \cite{lee2018propagating} &38.4 &- \\
  Pavllo et al. (243 frames) \cite{pavllo20193d} &37.2 &27.2\\ 
  Ours (243 frames) &\textbf{32.3} & \textbf{25.2}  \\\cline{1-3}

  \end{tabular}}%
  \end{threeparttable}
  
  \vspace{-1mm}
 \end{table}%
 
Moreover, our model sharply improves the lower bound of 3D pose estimation when using the ground-truth 2D keypoints as input. For this experiment, data augmentation is not applied as it can be regarded as using extra ground truth 2D keypoints. From Table~\ref{tab:result2}, the gap between our model and the baseline is nearly 5mm on Protocol 1. It indicates that if the performance of the bottom 2D keypoint detector is improved, our model can further boost the improvement.

Following \cite{pavllo20193d}, we report the model parameter number and an estimate of the floating-point operations (FLOPs) per frame at inference time to compare different models' computation complexity. The FLOPs are estimated in the same way as \cite{pavllo20193d}. Besides, to evaluate the models' inference efficiency, we report the inference frame per second (FPS) of different models by letting them estimate the 3D poses of a 10,000-frame test video frame by frame on a single
GeForce GTX 2080 Ti GPU. As shown in Table~\ref{tab:complex}, our 9-frame model holds similar computation complexity to the 243-frame baseline model \cite{pavllo20193d}. It achieves lower MPJPE with sharply fewer input frames, demonstrating the effectiveness of our proposed approach. On the other hand, even though the proposed model' inference speed is about 3.5 times lower than \cite{pavllo20193d}, they still hold an acceptable FPS, which is significantly higher than common 2D keypoint detection models \cite{fang2017rmpe,cao2019openpose}. As the complete 3D pose estimation process is a combination of 2D keypoint detection and 3D pose estimation, the inference speed of the proposed model will not be the bottleneck.

 \begin{table*}[htbp]
  \small\centering
  \caption{\label{tab:complex} Computational complexity, MPJPE, and frame per second (FPS) of different models under Protocol 1 on Human3.6. The computational complexity is computed without test-time augmentation used by all the models. The two numbers of ``$\approx$ FLOPs'' for our models represent the estimated floating-point operations of the bone direction prediction network and the bone length prediction network, respectively.}
  \begin{threeparttable}
  \scalebox{0.8}{
  \begin{tabular}{|c|c|c|c|c|}
  \cline{1-5}
  &Parameters &$\approx$ FLOPs &MPJPE & FPS\\\cline{1-5}
  Hossain \& Little \cite{rayat2018exploiting} ECCV'18 &16.96M &33.88M &58.3 & - \\
  Pavllo et al. \cite{pavllo20193d} (27 frames) CVPR'19 w/o dilation &29.53M &59.03M &49.3 & 486 \\
  Pavllo et al. \cite{pavllo20193d} (27 frames) CVPR'19 &8.56M &17.09M &48.8 & 1492 \\
  Pavllo et al. \cite{pavllo20193d} (81 frames) CVPR'19 &12.75M &25.48M &47.7 &1121 \\
  Pavllo et al. \cite{pavllo20193d} (243 frames) CVPR'19 &16.95M &33.87M &46.8 &863 \\ \cline{1-5}
  Ours (9 frames) &18.24M &29.97M + 4.58M &46.3 &759 \\
  Ours (27 frames) &31.88M &53.03M + 8.67M &45.3 &410 \\
  Ours (81 frames) &45.53M & 76.1M + 12.76M &44.6 &315 \\
  Ours (243 frames) &59.18M & 99.17M + 16.95M &44.1 &264
  \\\cline{1-5}

  \end{tabular}}%
  \end{threeparttable}
\end{table*}%

We further investigate the sensitivity of different hyper-parameters mentioned in Section~\ref{sec:id}. Overall, as shown in Table~\ref{tab:parameter}, the model's performance is insensitive to the setting of different hyper-parameters, indicating its strong robustness. Indeed, keeping increasing the sub-network number of the bone direction prediction network can still reduce the MPJPE. We set the final sub-network number to 2 as a good trade-off between the performance and the computational complexity.  

\begin{table}[htbp]
  \small\centering
  \caption{\label{tab:parameter} Parameter sensitivity test for our 81-frame model under Protocol 1 on Human3.6.}
  
  \begin{threeparttable}
  \scalebox{0.8}{
  \begin{tabular}{|c|c|c|c|c|c|c|}
  \cline{1-7}
  $\lambda_{D}$ &$\lambda_{L}$ &$\lambda_{JS}$ &$\gamma$ &$l$ &Num. of sub-network &MPJPE\\\cline{1-7}
  0.002 &0.05 &0.1 &10 &50&2&44.9 \\
  0.02 &0.05 &0.1 &10 &50&2&44.6\\
  0.2 &0.05 &0.1 &10 &50&2&44.6 \\
  0.02 &0.005 &0.1 &10 &50&2&45.0 \\
  0.02 &0.5 &0.1 &10 &50&2&44.6 \\
  0.02 &0.05 &0.01 &10 &50&2&45.0 \\
  0.02 &0.05 &1 &10 &50&2&44.9 \\ \cline{1-7}
  0.02 &0.05 &0.1 &1 &50&2&44.8 \\
  0.02 &0.05 &0.1 &10 &50&2&44.6 \\
  0.02 &0.05 &0.1 &100 &50&2&45.3 \\ \cline{1-7}
  0.02 &0.05 &0.1 &10 &10&2&45.0 \\
  0.02 &0.05 &0.1 &10 &50&2&44.6 \\
  0.02 &0.05 &0.1 &10 &100&2&44.6 \\ \cline{1-7}
  0.02 &0.05 &0.1 &10 &50&1&45.3 \\
  0.02 &0.05 &0.1 &10 &50&2&44.6 \\
  0.02 &0.05 &0.1 &10 &50&3&44.4 \\ \cline{1-7}
  
  \end{tabular}}%
  \end{threeparttable}
  \vspace{-1mm}
\end{table}%
\vspace{-1mm}

Table~\ref{tab:result3} shows the quantitative results of our full model and other baselines on MPI-INF-3DHP. Overall, MPI-INF-3DHP contains fewer training samples than human3.6M. This leads to better performance for the 81-frame models than the 243-frame models. Still, our model outperforms the baselines by a large margin. 

Overall, the comparison with the baseline model \cite{pavllo20193d} on Table~\ref{tab:result1},~\ref{tab:result2},~\ref{tab:result3} demonstrates that our superior performance is not only from the strong 2D keypoint detector, but also highly related to the proposed high-performance 3D model.

 \begin{table}[htbp]
  \small\centering
  \caption{\label{tab:result3} Quantitative comparisons of different models on MPI-INF-3DHP. }
  \begin{threeparttable}
  \scalebox{0.8}{
  \begin{tabular}{|c|c|c|c|}
  \cline{1-4}
  &PCK&AUC& MPJPE \\\cline{1-4}
  Mehta et al. \cite{mehta2017monocular} 3DV'17 &75.7&39.3&117.6 \\
  Mehta et al. \cite{mehta2017vnect} ACM ToG'17 &76.6&40.4&124.7  \\
  Pavllo et al. \cite{pavllo20193d} (81 frames) CVPR'19 &86.0&51.9&84.0 \\
  Pavllo et al. \cite{pavllo20193d} (243 frames) CVPR'19 &85.5&51.5&84.8 \\
  Lin et al. \cite{lin2019trajectory} BMVC'19 &82.4&49.6&81.9 \\\cline{1-4}
  Ours (81 frames) &\textbf{87.9}&\textbf{54.0}&\textbf{78.8}
  \\
  Ours (243 frames) &87.8&53.8&79.1
  \\\cline{1-4}

  \end{tabular}}%
  \end{threeparttable}
  
  \vspace{-1mm}
\end{table}%
\vspace{-1mm}

\subsection{Ablation Study} \label{sec:abl}

\begin{table}[t!]
  \centering\small
  \vspace{-1mm}
  \caption{\label{tab:ablation} Comparison of different models under Protocol 1 on Human3.6M. \textbf{Baseline} represents the baseline 81-frame model \cite{pavllo20193d}. Other rows represents the proposed anatomy-aware framework that decomposes the tasks into bone length prediction and bone direction prediction. \textbf{ML} refers to the use of MPJPE loss to solve the overfitting problem of the fully-connected residual network as Section~\ref{sec:length}. \textbf{BoneAtt} refers to the feeding of the bone length self-attention module for bone length re-weighting, instead of directly averaging the predicted bone lengths. \textbf{AUG} refers to the applying of data augmentation to solve the overfitting problem of the self-attention module. \textbf{JSL}, \textbf{LSC} and \textbf{ScoreAtt} refer to the applying of the joint shift loss, the incorporating of the fully-convolutional propagating architecture and the feeding of visibility scores by an implicit attention mechanism, respectively.}
  \begin{threeparttable}
  \scalebox{0.75}{
  \begin{tabular}{|c|cccccc|c|}
  \hline
    \cline{1-8}
    
    &\textbf{ML}&\textbf{BoneAtt}&\textbf{AUG}&\textbf{JSL}&\textbf{LSC}&\textbf{ScoreAtt}&MPJPE \\ 
    \cline{1-8}
    \textbf{Baseline}&\xmark&\xmark&\xmark&\xmark&\xmark&\xmark&47.7\\ 
    \textbf{Baseline}&\xmark&\xmark&\xmark&\cmark&\cmark&\cmark&46.4\\ 
    \cline{1-8}
    &\xmark&\xmark&\xmark&\xmark&\xmark&\xmark&48.4 \\
    &\cmark&\xmark&\xmark&\xmark&\xmark&\xmark&46.6 \\
    &\cmark&\cmark&\xmark&\xmark&\xmark&\xmark&46.5 \\
    \textbf{AF}&\cmark&\cmark&\cmark&\xmark&\xmark&\xmark&46.3 \\
    &\cmark&\cmark&\cmark&\cmark&\xmark&\xmark&45.8 \\
    &\cmark&\cmark&\cmark&\cmark&\cmark&\xmark&45.1 \\
    &\cmark&\cmark&\cmark&\cmark&\cmark&\cmark&44.6 \\
    \cline{1-8}
    \end{tabular}
    }
  \end{threeparttable}
  \vspace{1mm}
  \vspace{-1mm}
\end{table}
We next perform ablation experiments on Human3.6M under Protocol 1. For all the comparisons, we use the 81-frame models for the baseline \cite{pavllo20193d} and ours. They receive 81 consecutive frames as the input to predict the 3D joint locations and the bone directions of the current frame, respectively. 


We first show how each proposed module improves the model's performance in Table~\ref{tab:ablation}. For the very 
naive anatomy-aware framework, we adopt \textbf{Baseline} as the bone direction prediction network. We train the bone length prediction network and bone direction prediction network by the bone length loss and bone direction loss, respectively. We observe a drop of performance from 47.7mm to 48.4mm caused by the overfitting of the bone length prediction network. Once we handle the overfitting by applying the MPJPE loss and doing data augmentation which can be regarded as intrinsic parts of our anatomy-aware framework, MPJPE sharply drops to 46.3mm that is 1.4mm lower than \textbf{Baseline}. This demonstrates the framework's effectiveness. Moreover, we find that the joint shift loss, the fully-convolutional propagating architecture and the attentive feeding of visibility score reduce the error about 0.5mm, 0.7mm and 0.5mm, respectively. In the end, if we apply them to the baseline 81-frame work without the proposed anatomy-aware structure, the MPJPE increases to 46.4mm.

To further validate the proposed modules, the following models are compared to answer the following questions:

\begin{itemize}[leftmargin=0.2in]
\setlength\itemsep{0.0em}
\item
Q: Comparison between the proposed anatomy-aware network and the generic bone based representation.

M: \textbf{Baseline + Composition} 

A: Sun et al. \cite{sun2017compositional} propose compositional pose regression that transforms the task into a generic bone based representation. To demonstrate the effectiveness of the our anatomy-aware framework that decomposes the task explicitly into bone length and bone direction, we apply compositional pose regression on \textbf{Baseline}. Specifically, \textit{exactly} as \cite{sun2017compositional}, we modify \textbf{Baseline} to predict the bones and train it by the long range and short range joint shift loss (\emph{i.e.} compositional loss). The MPJPE is 47.4mm. Compared with our comparable anatomy-aware framework (without feeding \textbf{LSC} and \textbf{ScoreATT}) whose MPJPE is 45.8mm, this suggests that the explicit bone length and bone direction representation is more effective than the generic bone based representation because it can utilize global information across all the frames. It also makes better use of the relative joint shift supervision as our model obtains larger improvement from \textbf{JSL}.

\item 
Q: Comparison between our anatomy-aware network and directly imposing a bone length consistency loss.

M: \textbf{Baseline + ConsistencyLoss} 

A: A bone length consistency loss (as opposed to an explicit hard design) can be imposed in a straightforward manner across frames. To evaluate this idea, we further add a training loss term on \textbf{Baseline} to reduce the predicted bone length difference between randomly sampled frame pairs of a same video. The best MPJPE is 47.7mm. This indicates the uselessness of utilizing the bone length consistency by a form of training loss for this supervised learning task, compared with our solution.

\vspace{-0.00in}
\item 
Q: Whether the distant frames indeed help the prediction of bone length.

M: \textbf{AF(consecutive) + ML + BoneAtt + AUG + JSL} 

A: To validate this, we investigate the model that receives $l$ consecutive local frames as the input of the bone length prediction network for training/inference, with the current one in the center. As Section~\ref{sec:id}, $l$ is still set to 50 for both the training and inference process. The error increases from 45.8mm to 46.7mm. This demonstrates that randomly sampling frames from the entire video for bone length prediction indeed improves the model's performance, which is consistent with our motivation to decompose the task. 

\vspace{-0.00in}


\item 
Q: Whether the long skip connections and the independent training strategy indeed help the prediction of bone direction.

M: \textbf{AF + ML + BoneAtt + AUG + JSL + Baseline-D} 

A: It should be noticed that for the bone direction prediction network, \textbf{LSC} is two times deeper than \textbf{Baseline}. For fair comparison, we further design \textbf{Baseline-D} as the bone direction prediction network. Its structure is same as \textbf{LSC}, but with the long skip connections removed. Also, same as \textbf{Baseline}, the loss is only applied at the top and the back-propagation is not blocked between different sub-networks. The MPJPE is still 45.8mm. This indicates the uselessness of simply increasing the layer/parameter number of the temporal network.
\vspace{-0.00in}

\item 

Q: Whether utilizing implicit attention mechanism is an effective way to feed the visibility score feature.

M: \textbf{AF + ML + BoneAtt + AUG + JSL + LSC + Fusion}

A: We create a model for which we directly concatenate the visibility scores with the input 2D keypoints before feeding them into the network instead of utilizing implicit attention. The MPJPE is 44.8mm. It proves that the implicit attention mechanism provides a more effective way to feed the visibility score feature.

\end{itemize}

\subsection{Real-time 3D Pose Estimation}

\begin{figure}
\centering
\includegraphics[width=3.2in]{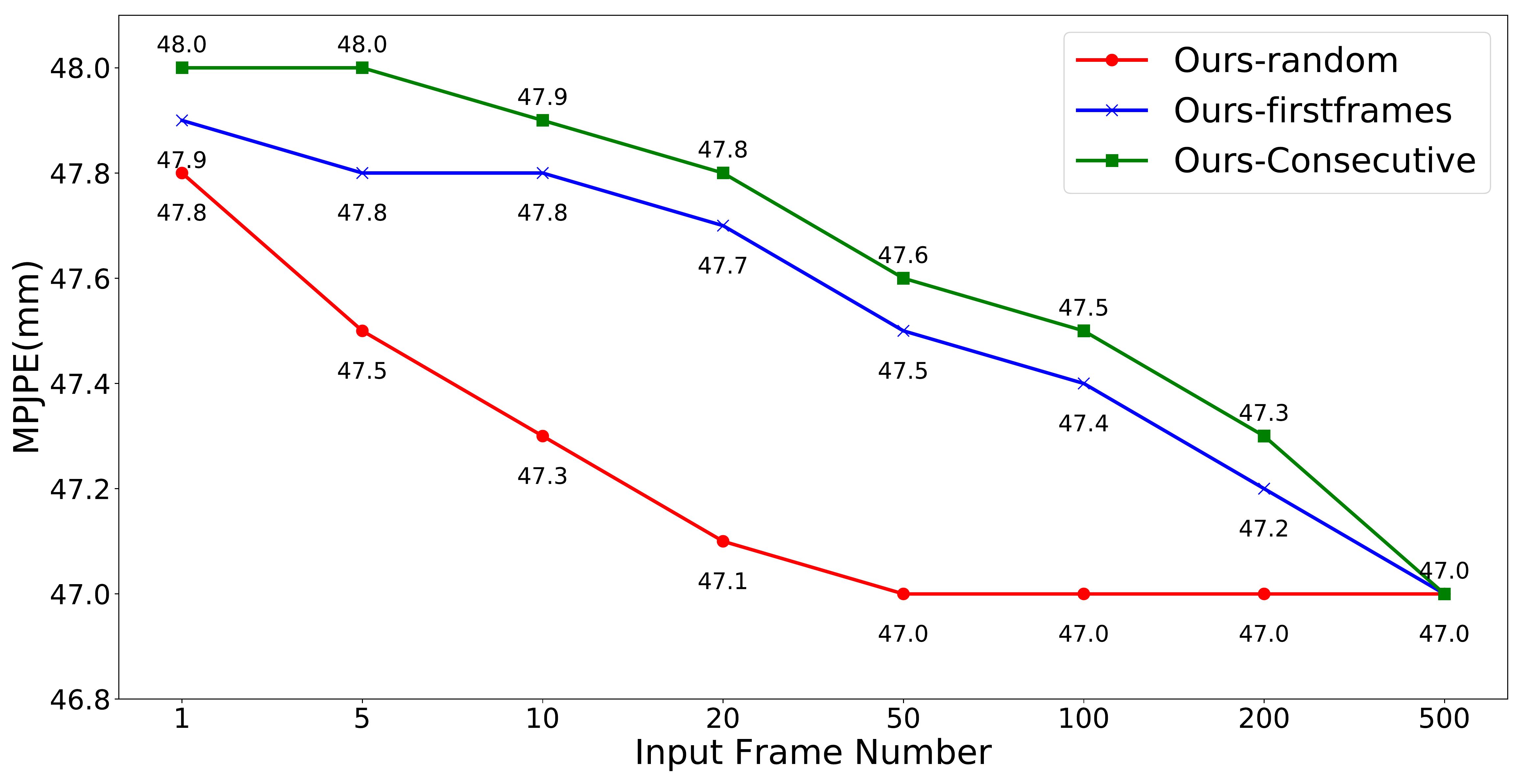}
\caption{The influence of the input frame number $M$ on different real-time models.}
\vspace{-1mm}
\label{fig:causal}

\end{figure}

As mentioned in Section~\ref{sec:expr}, our causal version model support real-time 3D human pose estimation. For the real-time 3D pose estimation, the inference speed is important and highly related to the frame selection strategy of the bone length prediction network. In this section, we compare different frame selection strategies for bone length prediction during the inference process. For all the comparisons,  same as Section~\ref{sec:abl}, we choose the 81-frame models for the baseline \cite{pavllo20193d} and ours. It should be noticed that the 3D pose estimation model can only leverage the information of the current and past frames.

\begin{itemize}
\setlength\itemsep{0.0em}
\item \textbf{BS-causal} refers to the baseline model with causal convolutions \cite{pavllo20193d}. To predict the 3D joint locations of the current frame, the input of the baseline model contains the 2D keypoint sequence of 81 consecutive frames with the current frame in the rightmost. The MPJPE is 49.8mm.


\item \textbf{Ours-random} refers to our real-time model that adopts the same frame selection strategy as our standard model presented in the main paper. For the bone direction prediction network, we input the 2D keypoint sequence of 81 consecutive frames as \textbf{BS-causal}. To predict the bone lengths of the $t$-th frame, we randomly sample $M$ frames before the $(t+1)$-th frame and input their 2D keypoints to the bone length prediction network. If $M$ is larger than $t$, we input the 2D keypoints of all the frames before the $(t+1)$-th frame.

\item \textbf{Ours-firstframe} refers to our real-time model that doesn't need to iteratively compute the bone lengths. Specifically, during the inference process, to predict the bone lengths of the $t$-th frame, we always select the 2D keypoints of the first $M$ frames of the video as input. In this situation, the model does not need to recalculate the bone lengths from the $(M+1)$-th frame. If $M$ is larger than $t$, we input the 2D keypoints of all the frames before the $(t+1)$-th frame. 

\item \textbf{Ours-consecutive} refers to our real-time model that receives consecutive local frames for bone length prediction. To predict the bone length of the $t$-th frame, we input the 2D keypoint sequence of $M$ consecutive frames with the $t$-th frame in the rightmost. Still, if $M$ is larger than $t$, we input the 2D keypoints of all the frames before the $(t+1)$-th frame.



\end{itemize}

\begin{table}[htbp]
  \small\centering
  \caption{\label{tab:causaltime} Relative average real-time inference speed on all the videos of Human3.6M test set.}
  \begin{threeparttable}
  \scalebox{0.8}{
  \begin{tabular}{|c|c|}
  \cline{1-2}
  & Relative Inference Speed \\\cline{1-2}
  BS-causal &1.0  \\
  Ours-random &0.28  \\
  Ours-firstframes & 0.46\\\cline{1-2}

  \end{tabular}}%
  \end{threeparttable}
  
\end{table}%
During the training process, for all the three models we propose (i.e. \textbf{Ours-random}, \textbf{Ours-firstframe}, \textbf{Ours-consecutive}), we still randomly sample 50 frames and input their 2D keypoints to the bone length prediction network. We don't observe further improvement of the model's performance when inputting the 2D keypoints of the first 50/50 consecutive local frames to train \textbf{Ours-firstframe}/\textbf{Ours-consecutive} as their inference process.

Fig.~\ref{fig:causal} shows how different settings of the input frame number $M$ for inference influences the models' performance. As we expect, \textbf{Ours-random} achieves best performance when $M$ is small. Moreover, randomly sampling 50 frames is sufficient for \textbf{Ours-random} to reach the best performance. However, it needs to predict the bone lengths frame by frame. On the other hand, \textbf{Ours-firstframe} achieves slightly better performance than \textbf{Ours-consecutive}. It indicates that predicting the bone lengths from the first frames of the video is more accurate than from consecutive local frames. We attribute it to the fact that the person's poses at the beginning of a video are commonly simpler on Human3.6M, this benefits the bone length prediction. To our surprise, even though we just update the bone lengths for the first 50 frames of each video and fix them from the $51$-th frame, \textbf{Ours-firstframes} still achieves great performance. The MPJPE is 47.5mm, which is 2.3mm lower than the baseline. More importantly, it is more efficient than \textbf{Ours-random}. Tables~\ref{tab:causaltime} shows different models' relative average inference speed on Human3.6M test set, $M$ is set to 50 for all the models. The inference speed of \textbf{Ours-firstframes} is about two times lower than \textbf{BS-causal}, because its bone direction prediction network that proposed in Section 3.2 of the main paper is nearly two times deeper than the baseline. However, \textbf{Ours-firstframes} is more efficient than \textbf{Ours-random} without updating the bone lengths frame by frame. 

\section{Conclusion}

We present a new solution to estimating the human 3D pose. Instead of directly regressing the 3D joint locations, we transform the task into predicting the bone lengths and directions. For bone length prediction, we make use of the frames across the entire video and propose an effective fully-connected residual network with a bone length re-weighting mechanism. For bone direction prediction, we add along skip connections into a fully-convolutional architecture for hierarchical prediction. Extensive experiments have demonstrated that the combination of bone length and bone direction is an effective intermediate representation to bridge the 2D keypoints and 3D joint locations. 

In recent years, as 3D human pose estimation has become a significant research topic for researchers to study, multiple directions are demonstrated to be promising for exploration. First, effective data augmentation algorithms \cite{cheng2019occlusion,li2020cascaded} are continuously proposed to guide the model to handle occluded or complex pose inputs. Moreover, the creation of generative adversarial network (GAN) \cite{goodfellow2014generative} enables a number of approaches \cite{yang20183d,wandt2019repnet} to utilize GAN for realistic and reasonable pose prediction, even in a weakly supervised setting. In addition, high-performance temporal models \cite{pavllo20193d,rayat2018exploiting} are created, which support very long 2D keypoint sequence as input and can adaptively capture significant information from keyframes. These directions are regarded as general directions since the proposed temporal models, adversarial training, and data augmentation algorithms can be generally applied to different research tasks other than 3D human pose estimation. In this paper, we focus on a more fundamental aspect of human pose estimation and create an effective learning representation for this task. We believe that exploring the human pose's learning representation is promising as the human body is a special Kinematic Tree-based structure different from other objects. The motion of the human body is drove by joint rotation with fixed bone lengths. We are delighted to see the human pose's learning representation evolved from the joint level to the bone vector level to the bone length/direction level that constantly improves human pose estimation. Based on our work, it may be illuminating for future works to keep exploring the relationship between the tasks of bone length prediction and bone direction prediction. Currently, we adopt two independent networks to predict bone directions and bone lengths. It is valuable to study whether the model can further improve the performance of each task by utilizing the knowledge captured from the other task. Applying the joint shift loss is one useful way. However, we believe that capturing this relationship at the network level as \cite{nam2017dual,lee2018stacked} for visual question answering and image-text matching will make extra improvement for accurate and smooth 3D human pose estimation in video.

\ifCLASSOPTIONcaptionsoff
  \newpage
\fi

\bibliographystyle{IEEEtran}
\bibliography{ref.bib}

\end{document}